\documentclass[letterpaper]{article} 
\usepackage{aaai25}  
\usepackage{times}  
\usepackage{helvet}  
\usepackage{courier}  
\usepackage[hyphens]{url}  
\usepackage{graphicx} 
\usepackage{natbib}  
\usepackage{caption} 
\usepackage{xcolor}

\usepackage{microtype}
\usepackage{url}
\usepackage{booktabs}
\usepackage{makecell}
\usepackage{times}
\usepackage{latexsym}
\usepackage{amsmath}
\usepackage{graphicx}
\usepackage{subfigure}
\usepackage{fontawesome}
\usepackage{bigdelim}
\usepackage{pifont}
\usepackage{inconsolata}
\usepackage[T1]{fontenc}
\usepackage[usestackEOL]{stackengine}
\usepackage[utf8]{inputenc}
\usepackage{multirow}

\usepackage{microtype}
\usepackage{caption}
\usepackage{subcaption}
\usepackage{arydshln} 
\newcommand*\samethanks[1][\value{footnote}]{\footnotemark[#1]}
\newcommand{\modelname}{\textsc{ReasonEval}}

\title{Evaluating Mathematical Reasoning Beyond Accuracy}

\author{
    Written by AAAI Press Staff\textsuperscript{\rm 1}\thanks{With help from the AAAI Publications Committee.}\\
    AAAI Style Contributions by Pater Patel Schneider,
    Sunil Issar,\\
    J. Scott Penberthy,
    George Ferguson,
    Hans Guesgen,
    Francisco Cruz\equalcontrib,
    Marc Pujol-Gonzalez\equalcontrib
}

\author{Shijie Xia\textsuperscript{\rm 1,2,5}, Xuefeng Li\textsuperscript{\rm 1,2,5}, Yixin Liu\textsuperscript{\rm 4}, Tongshuang Wu\textsuperscript{\rm 3}\thanks{\,\, Corresponding Authors.}, Pengfei Liu\textsuperscript{\rm 1,2,5}\samethanks }

\affiliations{

   \textsuperscript{\rm 1} Shanghai Jiao Tong University, 
   \textsuperscript{\rm 2} Shanghai Artificial Intelligence Laboratory\\
   \textsuperscript{\rm 3} Carnegie Mellon University,
      \textsuperscript{\rm 4} Yale University,
    \textsuperscript{\rm 5} Generative AI Research Lab (GAIR)\\
    \{shijiexia, xuefengli, pengfei\}@sjtu.edu.cn, yixin.liu@yale.edu, sherryw@cs.cmu.edu
    }

\begin{document}
\maketitle
\begin{abstract}
The leaderboard of Large Language Models (LLMs) in mathematical tasks has been continuously updated. However, the majority of evaluations focus solely on the final results, neglecting the quality of the intermediate steps. This oversight can mask underlying problems, such as logical errors or unnecessary steps in the reasoning process. To measure reasoning beyond final-answer accuracy, we introduce \modelname, a new methodology for evaluating the quality of reasoning steps. \modelname\ employs \textit{validity} and \textit{redundancy} to characterize the reasoning quality, as well as accompanying LLMs to assess them automatically. We explore different design options for the LLM-based evaluators and empirically demonstrate that \modelname, when instantiated with base models possessing strong mathematical knowledge and trained with high-quality labeled data, consistently outperforms baseline methods in the meta-evaluation datasets. We also highlight the strong generalization capabilities of \modelname. By utilizing \modelname\ to evaluate LLMs specialized in math, we find that an increase in final-answer accuracy does not necessarily guarantee an improvement in the overall quality of the reasoning steps for challenging mathematical problems. Additionally, we observe that \modelname\ can play a significant role in data selection. We open-source the best-performing model, meta-evaluation script, and all evaluation results to facilitate future research.
\end{abstract}

\begin{links}
  \link{Code}{https://github.com/GAIR-NLP/ReasonEval}
\end{links}
\section{Introduction}

Mathematical reasoning, a core cognitive skill, is crucial for resolving complex problems and making informed decisions~\citep{hendrycks2021measuring,lewkowycz2022solving}, playing a significant role in large language models (LLMs) research~\citep{azerbayev2023llemma,lu2023survey}.
Given its significance, reliably evaluating mathematical reasoning in LLMs becomes crucial.
Current methodologies to evaluate mathematical reasoning in LLMs focus primarily on the final result~\citep{luo2023wizardmath,abel,yu2023metamath}, neglecting the intricacies of the reasoning process.
For example, the OpenLLM leaderboard,\footnote{\url{https://huggingface.co/spaces/HuggingFaceH4/open_llm_leaderboard}} a relatively well-recognized benchmark for LLMs, uses overall accuracy to assess models' mathematical reasoning. Despite being effective to some degree, such evaluation practice could mask underlying issues such as logical errors or unnecessary steps that compromise accuracy and efficiency of reasoning steps. In this work, we argue that a desirable evaluation criterion for mathematical reasoning
encompasses not only the accuracy of the final answer but also the correctness and efficiency of each step in the reasoning process.
Moreover, it is imperative that the evaluation metrics be open-source and replicable to ensure transparency and reliability. 

\begin{figure*}[!t]
\centering 
    \includegraphics[width=0.9\textwidth]{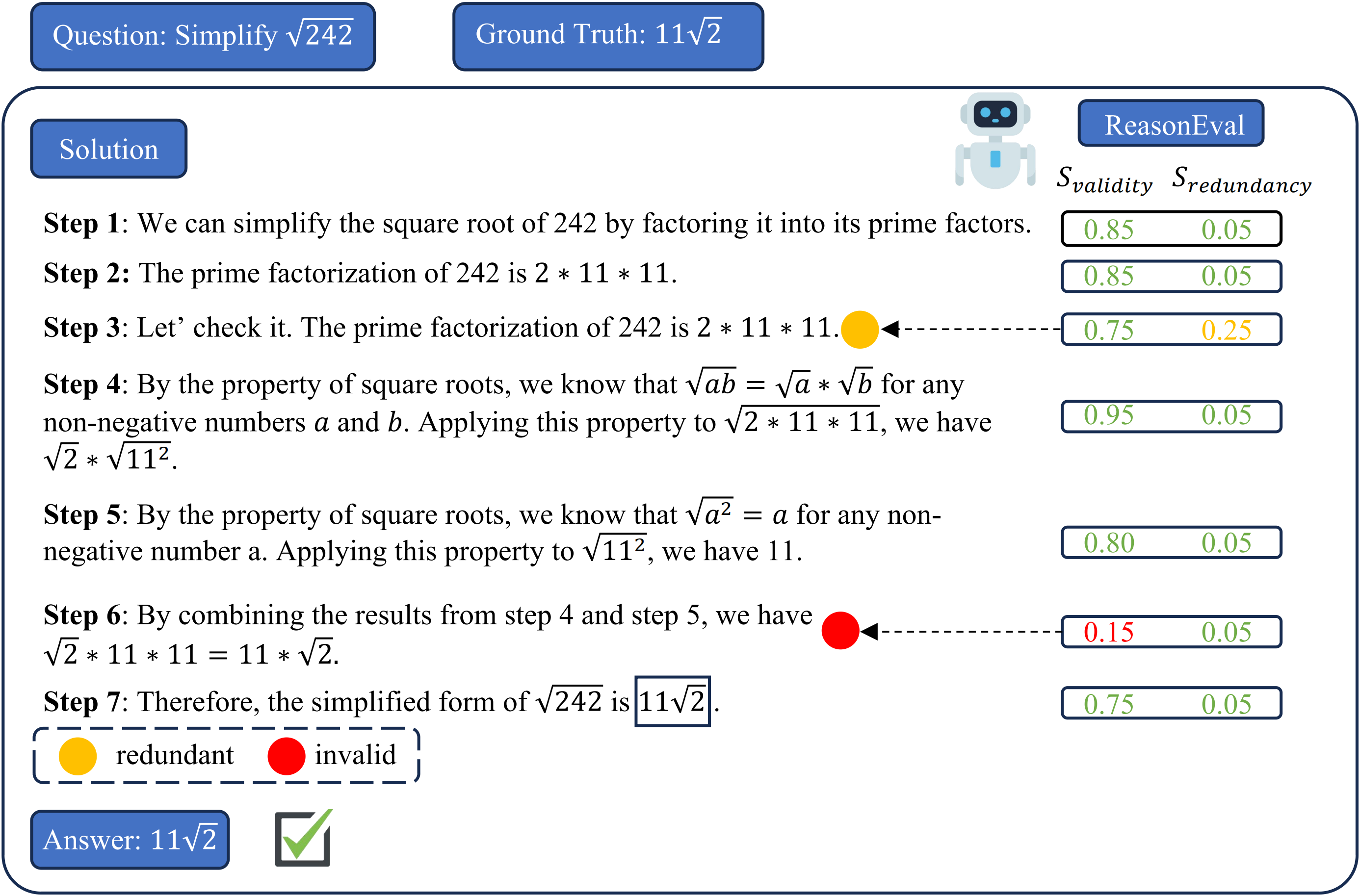}
\caption{Given a solution to a math problem, \modelname\ scores each step and identifies the potential error location, serving as an extension to verify the final answer only.}
\label{introduction}
\end{figure*}

Our design philosophy stems from the fact that a correct final answer does not guarantee a flawless reasoning process~\citep{lewkowycz2022solving,uesato2022solving}, and excessive or irrelevant reasoning steps can lead to potential errors as well as increased computational costs~\citep{zhang2023interpretable}. 
Once these issues go unnoticed, they can cause problems in many application scenarios. For example, in K12 mathematics education, incorrect or redundant solution steps provided by LLMs could mislead students. There have been some recent works related to the above evaluation principle. Specifically, ~\citet{uesato2022solving,lightman2023let,wang2023math} present process reward models for mathematical reasoning,\footnote{~\citet{uesato2022solving} and~\citet{lightman2023let} are close-source.} which focus on their utility as verifiers~\citep{cobbe2021training} to boost the accuracy (i.e., by generating many candidate solutions and selecting the one ranked highest by the verifier), leaving its underlying potential to identify reasoning errors\footnote{Identifying reasoning errors and ranking reasoning steps are different tasks. As also highlighted by~\citet{tyen2023llms}, even state-of-the-art (SOTA) models like GPT-4~\citep{openai2023gpt} still struggle to detect errors in reasoning tasks.} and redundancy.~\citet{clinciu2021study,golovneva2022roscoe} rely on embedding-based methods to measure the quality of general reasoning explanation, which has limitations in handling diverse reasoning steps from various models.

In response to these challenges, we propose \modelname, a suite comprising a new evaluation methodology with defined metrics for assessing mathematical reasoning quality and corresponding LLM-based evaluators for automated calculation. As illustrated in Figure~\ref{introduction}, \modelname\ emphasizes the \emph{validity} (i.e., the step contains no mistakes in calculation and logic) of each reasoning step, and evaluates the \emph{redundancy} (i.e., the step lacks utility in solving the problem but is still valid) of steps to ensure efficiency. We explore different design options for the LLM-based evaluators and show training such a high-quality evaluator for mathematical reasoning is not a straightforward task, which requires an LLM with strong mathematical knowledge as well as high-quality training data. We empirically demonstrate that \modelname\ using the optimal design can consistently outperform baseline methods by a large margin in the meta-evaluation datasets~(\S\ref{results}). Moreover, we highlight the strong generalization ability of \modelname\ (\S\ref{ood}).

Furthermore, we show the utility of \modelname\
through the lens of two preliminary applications, as detailed in \S\ref{section6}: (1) evaluating different LLMs trained for mathematical reasoning; and (2) selecting high-quality data to train such LLMs. We have the following main findings:
(1) We find that an improvement in the result accuracy is not sufficient to ensure an enhancement in the overall quality of reasoning steps in challenging mathematical problems;
(2) We find that the model scale, the base model, and the training methods have significantly influenced the quality of reasoning steps;
(3) We find that \modelname\ can select high-quality training data to improve the efficiency of solving problems and the quality of solutions.
These findings pave the way for future work to design methods that take into account the process of problem-solving.

Overall, our main contributions are as follows:

(1) We recognize the potential gap between what we are evaluating and what we are desiring in mathematical reasoning, prioritize validity and redundancy aspect to address the misalignments and inefficiencies that can arise in mathematical reasoning processes (\S\ref{Section3}).
(2) We design a meta-evaluation testbed to assess the reliability (\S\ref{experiment}) and utility (\S\ref{section6}) of mentioned metrics, guiding us to a superior metric design, solidifying the foundation for future exploration and significantly lowering trial and error costs.
(3) We open-source our best-performing metric, meta-evaluation script and all evaluation results to facilitate future research.

\section{Preliminaries}\label{section2}

\subsection{Problem Formulation}

Given a mathematical problem $q$, solution steps $\hat{\mathbf{{h}}} = \{\hat{h}_1, . . . , \hat{h}_N \}$ and an answer $\hat{a}$ generated by LLMs, the goal is to evaluate how well the generated solution and answer are. Usually, the ground truth answer $a$ is available but the reference solution step $\mathbf{{h}} = \{{h}_1, . . . , {h}_M \}$ is not always provided.

\subsection{Existing Evaluation Methodology}

\paragraph{Answer-based Matching}
Most of the existing works~\citep{luo2023wizardmath,abel,yu2023metamath} measure the mathematical reasoning quality by directly comparing the final answer (i.e., $\hat{a}$ and $a$) and calculating the overall accuracy on a given dataset.

\paragraph{Reference-based Scoring}
Instead of only using the final result as a scoring criterion, some works~\citep{sawada2023arb} try to measure the reasoning quality by comparing the similarity between generated and reference solution steps (i.e., $\hat{\mathbf{{h}}}$ and $\mathbf{{h}}$). Although datasets like \texttt{MATH}~\citep{hendrycks2021measuring} and \texttt{GSM8K}~\citep{cobbe2021training} provide the ground truth solutions, 
the existence of diverse reasoning paths leading to the same answer $a$ renders reliance on any single one of them as a reference unreliable.

\paragraph{Prompting-based Method} 

This method directly asks LLMs with a well-defined prompt to generate a judgment for a given generated solution and answer~\citep{hao2024llmreasonersnewevaluation}. This approach usually requires a very powerful LLM, often GPT-4, which may not be practical for iterative model development due to cost and transparency concerns.

The aforementioned methods either focus solely on the final results (e.g., $a$) or are limited by the use of reference solutions and proprietary LLMs, and most of them only concentrate on evaluating ``correctness'', neglecting other aspects such as the redundancy of solution steps. This has inspired us to propose a set of evaluation criteria that better aligns with the reasoning models in the era of LLMs.

\section{\modelname: Metrics and Implementations} \label{Section3}

\subsection{Design Principle}
We argue that for tasks involving multi-step reasoning, measuring the quality of reasoning should not solely depend on the correctness of the final result. It should also consider (1)
the accuracy of each reasoning step;
(2) the efficiency of the reasoning process. 
To this end, we design evaluators that assess reasoning steps regarding validity and redundancy in a step-by-step format, checking whether each step advances well towards solving the problem.
Precisely, \textbf{validity} denotes the step contains no mistakes in calculation and logic, and \textbf{redundancy} describes the step lacks utility in solving the problem but is still valid.

\subsection{Scoring Scheme} \label{Task Formulation}

Following this principle, we formulate such a metric design process as a classification task. We classify each step into three classes: positive, neutral, and negative. The positive label indicates that the step is correct and contributes to solving the question, the neutral label represents that the step is correct but does not make any progress, and the negative label signifies an incorrect step. Given a question $q$ and the solution steps $\hat{\mathbf{{h}}} = \{\hat{h}_1, . . . , \hat{h}_N \}$, each step will be assigned the probability of the three classes as follows:
\begin{align}
   \{{p}^{1}, . . . , {p}^{N} \} &= \modelname(q, \hat{h}_1, . . . , \hat{h}_N)\label{eq:reasoneval} \\
   p^i &= (p^i_{\text{positive}}, p^i_{\text{neutral}}, p^i_{\text{negative}})
\end{align}

The validity score, concerning only the correctness of the reasoning steps, is defined as:
\begin{align}
    S_{\text{validity}}^i &= p^i_{\text{positive}} + p^i_{\text{neutral}}
\end{align}

The redundancy score, concerning the utility of the steps given that they are valid, is defined as:
\begin{align}
    S_{\text{redundancy}}^i &= p^i_{\text{neutral}}
\end{align}

We can aggregate the \textbf{step-level} scores to get the \textbf{solution-level} scores. We use the `min' and `max' operations:
\begin{align}
    S_{\text{validity}}^{\text{all}} &= \min(S^1_{\text{validity}}, . . . , S^N_{\text{validity}})\\
    S_{\text{redundancy}}^{\text{all}} &= \max(S^1_{\text{redundancy}}, . . . , S^N_{\text{redundancy}}) 
\end{align}

We describe the detailed justification of our scoring scheme in Appendix \ref{scoring_scheme}.

\subsection{Model Architecture}

\paragraph{LLM Backbone}

Our defined evaluation method (Eq.\ref{eq:reasoneval}) can be implemented by directly prompting existing LLMs or fine-tuning LLMs using supervised dataset. To make a comprehensive study, in this work we instantiate \modelname\ by different types of LLMs, which vary in base model types (e.g., Llama2~\citep{touvron2023llama} and Mistral~\citep{jiang2023mistral}), model sizes (e.g., 7B, 34B), and post-training strategies (e.g., continued pretraining~\citep{azerbayev2023llemma} and fine-tuning).

\paragraph{Task-specific Layer}
Our evaluator's architecture is identical to the base model, except that the linear layer for next-token prediction is replaced with a linear layer for outputting the possibilities of each class.
After normalization with the softmax layer, we get the possibility for each class.

\subsection{Fine-tuning}
The above task formulation allows us to utilize the existing dataset, \texttt{PRM800K}~\citep{lightman2023let}, as training data in \modelname.
\texttt{PRM800K} contains about 800,000 step-level labels over 75,000 solutions. It is collected by employing humans to label the step-by-step solutions generated by GPT-4 to \texttt{MATH} problems. The label categories for the reasoning steps are identical to those mentioned in \S\ref{Task Formulation}. In the training phase, the loss function only includes the last token of each step, as it contains the full information about the step. We also take the last token of each step for prediction at test time.
We provide the details on training and splitting solutions in Appendix \ref{Training_details}.

\section{Meta Evaluation} \label{experiment}

\subsection{Meta-evaluation Setup} \label{setup}

\begin{table*}[htbp]
  \centering
    \begin{tabular}{lcccccccc}
    \toprule
     & \multicolumn{4}{c}{MR-MATH-invalid} & \multicolumn{4}{c}{MR-MATH-redundant} \\
   \cmidrule(lr){2-5} \cmidrule(lr){6-9}
          & \multicolumn{2}{c}{\emph{Solution-level}} & \multicolumn{2}{c}{\emph{Step-level}} & \multicolumn{2}{c}{\emph{Solution-level}} & \multicolumn{2}{c}{\emph{Step-level}} \\
     \cmidrule(lr){2-3} \cmidrule(lr){4-5} \cmidrule(lr){6-7} \cmidrule(lr){8-9}
          & \textbf{F1 Score} & \textbf{AUC} & \textbf{F1 Score} & \textbf{AUC} & \textbf{F1 Score} & \textbf{AUC} & \textbf{F1 Score} & \textbf{AUC} \\
    \midrule
  \multicolumn{9}{l}{\textbf{\emph{Embedding-based Methods}}} \\
    \midrule
    ROSCOE-SA & 48.2  & 57.5  & -     & -     & 50.7  & 53.9  & -     & - \\
    ROSCOE-SS & 51.6  & 49.6  & -     & -     & 52.0  & 52.7  & -     & - \\
    \midrule
  \multicolumn{9}{l}{\textbf{\emph{Prompting-based Methods}}} \\
    \midrule
    GPT-3.5-turbo & 59.7  & -     & 53.2  & -     & 53.0  & -     & 51.5  & - \\
    GPT-4 & 73.2  & -     & 61.0  & -     & 57.1  & -     & 54.2  & - \\
    \midrule
   \multicolumn{9}{l}{\textbf{\emph{Step-level Evaluators}}} \\
    \midrule
  Math-shepherd-Mistral-7b &70.1  & 77.3  & 60.0  & 77.2  & 50.4  & 54.5  & 42.7  & 53.0 \\
  \hdashline
\modelname$_{\text{Llama2-7B}}$  & 66.7  & 79.5  & 60.8  & 80.0  & 60.4  & 62.8  & 59.0  & 68.6  \\
 \modelname$_{\text{WizardMath-7B-V1.0}}$& 72.8  & 81.9  & 67.7  & 83.9  & 60.5  & \textbf{65.6} & 59.0  & 68.3  \\
\modelname$_{\text{Mistral-7B}}$ & 78.0  & 85.1  & 68.6  & 85.7  & 60.7  & 63.4  & 59.7  & 70.9  \\
\modelname$_{\text{Llemma-7B}}$ & 74.7  & 84.3  & 76.6  & 90.5  & 59.6  & 63.0  & 58.6  & 68.3  \\
\modelname$_{\text{Abel-7B-002}}$ & 77.3  & 86.2  & 70.4  & 90.5  & 58.6  & 63.6  & 59.5  & 71.8  \\
\modelname$_{\text{WizardMath-7B-V1.1}}$ & 78.6  & 87.5  & 73.9  & 89.5  & \textbf{61.6} & 64.8  & \textbf{59.7} & \textbf{72.2} \\
\modelname$_{\text{Llemma-34B}}$ & \textbf{79.6} & \textbf{90.8} & \textbf{77.5} & \textbf{92.8} & 58.3  & 62.7  & 57.5  & 67.3  \\
    \bottomrule
    \end{tabular}
    \caption{Comparison of methods for automatic evaluation of reasoning steps in \texttt{MR-MATH}. For any methods that require setting a threshold, we report the Area Under the Curve (AUC) metric.}
  \label{error detector}%
\end{table*}%

\paragraph{Meta-evaluation\footnote{Meta-evaluation refers to evaluating the performance of evaluators themselves.} Datasets} 

Inspired by \citet{zeng2024mrgsm8k}, which propose a benchmark named Meta-Reasoning-GSM8K (\texttt{MR-GSM8K}), we construct a meta-evaluation dataset \texttt{MR-MATH} to better assess how well different evaluators can detect errors in reasoning steps. It is constructed as follows: (1) To collect the first type of errors affecting the correctness of steps, we recruit undergraduates who have a solid mathematical background to label solutions generated by Abel~\citep{abel} and WizardMath~\citep{luo2023wizardmath}. We collect 83 samples with incorrect steps and 76 without, pinpointing the first error location in the former. All the solutions reach the correct final answers, meaning the evaluators need to judge correctness based on the process rather than the outcome. Additionally, since the style of solutions differs from the training set of \texttt{PRM800K} (See examples of solutions in Appendix \ref{appendix:lanuguage style}), it tests the generalization of \modelname\ to some degree. (2) For the second type of errors affecting the efficiency of problem solving process, as they are more rarer than the first one, we sample solutions from the test set of \texttt{PRM800K} directly, containing 150 samples with redundant steps and 150 samples without.

\begin{table*}[htbp]
  \centering
    \begin{tabular}{lccccccccc}
    \toprule
          &       & \multicolumn{4}{c}{MR-GSM8K-original}  & \multicolumn{4}{c}{MR-GSM8K-reversed} \\
          \cmidrule(lr){3-6} \cmidrule(lr){7-10}
          &       & \multicolumn{2}{c}{\textit{Solution-level}} & \multicolumn{2}{c}{\textit{Step-level}} & \multicolumn{2}{c}{\textit{Solution-level}} & \multicolumn{2}{c}{\textit{Step-level}} \\
            \cmidrule(lr){3-4} \cmidrule(lr){5-6} \cmidrule(lr){7-8} \cmidrule(lr){9-10}
          & \textbf{OOD}   & \textbf{F1 Score} & \textbf{AUC}   & \textbf{F1 Score} & \textbf{AUC}   & \textbf{F1 Score} & \textbf{AUC}   & \textbf{F1 Score} & \textbf{AUC} \\
    \midrule
    \textit{\textbf{Embedding-based Methods}} &       &       &       &       &       &       &       &       &  \\
    \midrule
    ROSCOE-SA & \ding{51}     & 51.6  & 54.4  & -     & -     & 54.5  & 57.9  & -     & - \\
    ROSCOE-SS & \ding{51}     & 49.6  & 60.1  & -     & -     & 49.6  & 52.1  & -     & - \\
    \midrule
    \textit{\textbf{Prompting-based Methods}} &       &       &       &       &       &       &       &       &  \\
    \midrule
    GPT-3.5-turbo & -     & 54.9  & -     & 52.3  & -     & 54.3  & -     & 49.9  & - \\
    GPT-4 & -     & 81.7  & -     & 69.0  & -     & 72.2  & -     & 52.2  & - \\
    \midrule
    \textit{\textbf{Step-level Evaluators}} &       &       &       &       &       &       &       &       &  \\
    \midrule
    Math-shepherd-Mistral-7b & \ding{55}     & \textbf{86.0} & \textbf{93.9} & 73.4  & 88.5  & \textbf{77.2} & \textbf{88.0} & 59.6  & 77.9  \\
    \hdashline

    \modelname$_{\text{Mistral-7B}}$ & \ding{51}     & 61.8  & 79.8  & 62.9  & 86.1  & 61.0  & 71.9  & 61.5  & 84.3  \\

    \modelname$_{\text{WizardMath-7B-V1.1}}$ & \ding{51}     & 74.1  & 90.7  & 72.8  & \textbf{91.4} & 74.4  & 86.3  & \textbf{70.5} & \textbf{90.5} \\
    \modelname$_{\text{Llemma-34B}}$ & \ding{51}     & 81.0  & 88.1  & \textbf{73.5} & 86.8  & 76.1  & 84.1  & 69.3  & 85.0  \\
    \bottomrule
    \end{tabular}%
    \caption{Comparison of methods for automatic evaluation of reasoning steps in \texttt{MR-GSM8K}. ``OOD'' represents that its training data contains the labeled solutions for problems of the dataset. The results of prompting-based methods are from~\citet{zeng2024mrgsm8k}.}
      \label{ood_results}%
\end{table*}%

\paragraph{Evaluators} We compare three methods to evaluate reasoning steps automatically: embedding-based methods, prompting-based methods and step-level evaluators. For the embedding-based methods, we choose ROSCOE ~\citep{golovneva2022roscoe}, a SOTA method among them. Specifically, we select the semantic alignment metrics (ROSCOE-SA) and the semantic similarity metrics (ROSCOE-SS), which do not require references solution steps. For the prompting-based methods, we select GPT-3.5-turbo and GPT-4, two mainstream generation models. We use the prompt suggested by~\citet{zeng2024mrgsm8k} for the invalid errors and the modified version for the redundant errors. For the step-level evaluators, we select 7 base models for \modelname: Llama-2-7B, Mistral-7B, Llemma-7B, Llemma-34B, WizardMath-7B-V1.0~\citep{luo2023wizardmath}, WizardMath-7B-V1.1, Abel-7B-002~\citep{abel}. The llemma series is continuing pertaining on math-based corpus. The Abel and WizardMath series are fine-tuning for solving mathematical problems. We also select the open-source SOTA process reward model Math-shepherd-Mistral-7B~\citep{wang2023math} to compare. 
The settings for these methods are in Appendix \ref{settings_for_meta_evalautions}.

\paragraph{Meta-evaluation Metrics} We test two abilities: judging whether the solution contains errors and locating the error step. In the first task the ground truth is labels for solutions (solution-level), and in the second tasks the ground truth is labels for steps (step-level). Since both are classification tasks, we present the macro F1 score as a performance metric. Additionally, for any methods that require setting a threshold, we report the Area Under the Curve (AUC) metric, which is a more balanced evaluation of performance across different threshold settings.

\subsection{Results and Analysis} \label{results}

\paragraph{Overall Results}We present our main results in Table \ref{error detector}. \modelname\ outperforms baseline methods across all error types and label levels. It is noteworthy that the identification of errors at the step level is generally more challenging than at the solution level for all methods. This suggests a higher complexity in pinpointing
errors at the individual step level. Furthermore, identifying redundant errors is more intricate than invalid errors due to the inherent ambiguity involved in the former.

\begin{figure}[htbp]
\centering 
    \includegraphics[width=0.45\textwidth]{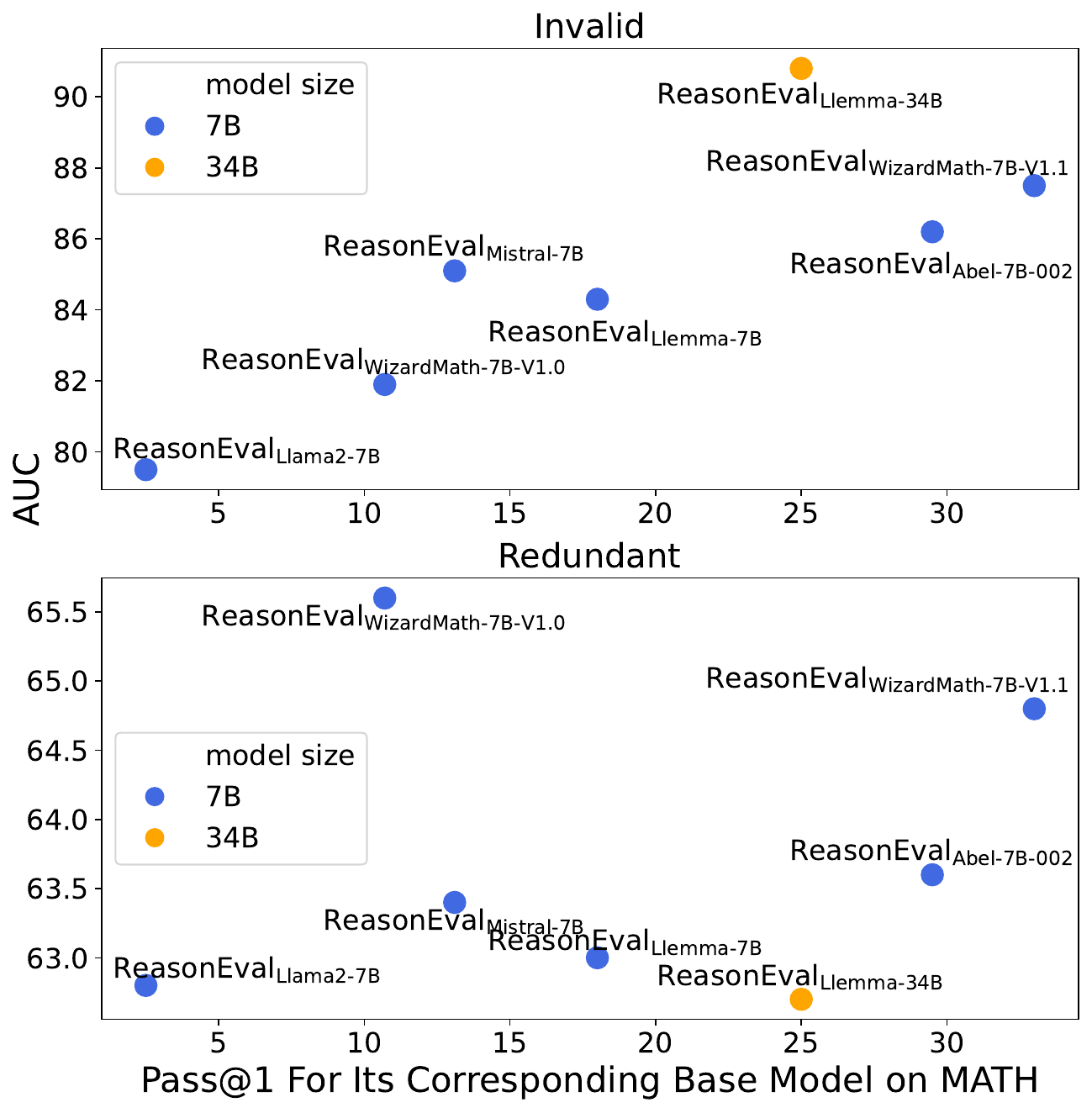}
\caption{Correlation between Pass@1 for the base model on \texttt{MATH} and AUC for the solution-level labels.}
\label{base model correlation}
\end{figure}

\paragraph{Analysis on Base Models} We investigate the correlation between the base LLMs' mathematical reasoning capabilities and the performance of \modelname\ fine-tuned from them. In Figure \ref{base model correlation}, for the invalid errors, an increase in AUC shows a positive correlation with the base models' ability to solve \texttt{MATH} problems, indicating that enhancing mathematical problem-solving abilities is beneficial for identifying invalid errors. It is noted that the Llemma-34B outperforms all 7B models, although its Pass@1 is not the highest. It highlights the importance of model scale and the continued pretraining over math-related corpus. For the redundant errors, the distinction across different base models is small, and it does not show the correlation as the invalid errors. This may be due to developers for these models prioritizing the correctness of solutions over their efficiency, resulting in similar performances in this aspect across various LLMs.

\paragraph{Analysis on Training Data} In Table \ref{error detector}, the Mistral-7B trained on \texttt{Math-Shepherd} falls behind that trained on \texttt{PRM800K}. We summarize the information about \texttt{Math-shepherd} and \texttt{PRM800K} in Table \ref{dataset_comparison}. For \texttt{Math-shepherd}, although having 6x training data, there is more noise in the step-level labels generated with the unsupervised method, thus harming its precision in identifying errors. And the only two classes of labels also make it limited in evaluating redundancy. Nonetheless, the data synthesis method is promising. We leave it for future work to optimize this method to reduce noise.
\begin{table}[htbp]
  \centering
    \begin{tabular}{lcccc}
    \toprule
    \textbf{Dataset} & \textbf{\#S} &  \textbf{\#C} & \makecell[l]{\textbf{Problem}\\\textbf{Source}} & \makecell[l]{\textbf{Human}\\\textbf{Ann.}} \\
    \midrule
    Math-shepherd & \text{440K} &\makecell[l]{2} & \makecell[l]{MATH,\\GSM8K}  & \ding{55} \\
    \midrule
    PRM800K & \text{75K}   & \makecell[l]{3} & MATH & \ding{51} \\
    \bottomrule
    \end{tabular}
      \caption{Comparison between \texttt{Math-shepherd} and \texttt{PRM800K}. ``\#S'' represents the number of labeled solutions. ``\#C'' represents the number of label categories for the reasoning steps. For \texttt{Math-shepherd}, the categories are ``correct" and ``incorrect". ``Human Ann.'' indicates whether the labels for the reasoning steps are generated by human annotations.}
  \label{dataset_comparison}%
\end{table}

\subsection{Out-Of-Distribution Generalization} \label{ood}
\paragraph{Setup} To assess the out-of-distribution generalization of $\modelname_{}$, we evaluate its performance on \texttt{MR-GSM8K}~\citep{zeng2024mrgsm8k}. We select two distinct types of questions from the dataset. The first type of question is from the original GSM8K. The second type, termed reversed reasoning~\citep{yu2023metamath}, involves concealing one of the inputs and requiring the computation of the missing input using the provided original answer. Both problem sets cover the full test set of GSM8K, comprising approximately 1.4K samples each. The step-by-step solutions for these questions are sampled from MetaMath-7B~\citep{yu2023metamath} and include human annotations labeling the correctness of each step. Since the training data of $\modelname_{}$ does not include the labeled solutions for these types of problems, this evaluation serves as a robust test of its generalization ability. All settings are identical to those described in \S\ref{setup}.
\paragraph{Results} We present the results in Table \ref{ood_results}. We observe that both $\modelname_{\text{Llemma-34B}}$ and $\modelname_{\text{WizardMath-7B-V1.1}}$ achieve superior performance at the step level and approach the performance of GPT-4 at the solution level, demonstrating strong generalization capabilities. Additionally, $\modelname_{}$ is more robust to the question types compared to other methods. For Math-shepherd-Mistral-7B, it performs best in solution-level labels but lags behind in step-level labels, suggesting that the noise in step labels from \texttt{Math-Shepherd} negatively impacts its ability to accurately locate error steps.

\section{Utilizing \modelname\ for Evaluating and Improving Reasoning Quality} \label{section6}

\subsection{Evaluating Reasoning Quality of LLMs Specialized in Math}

\begin{figure}[htbp]
\centering 
    \includegraphics[width=0.45\textwidth]{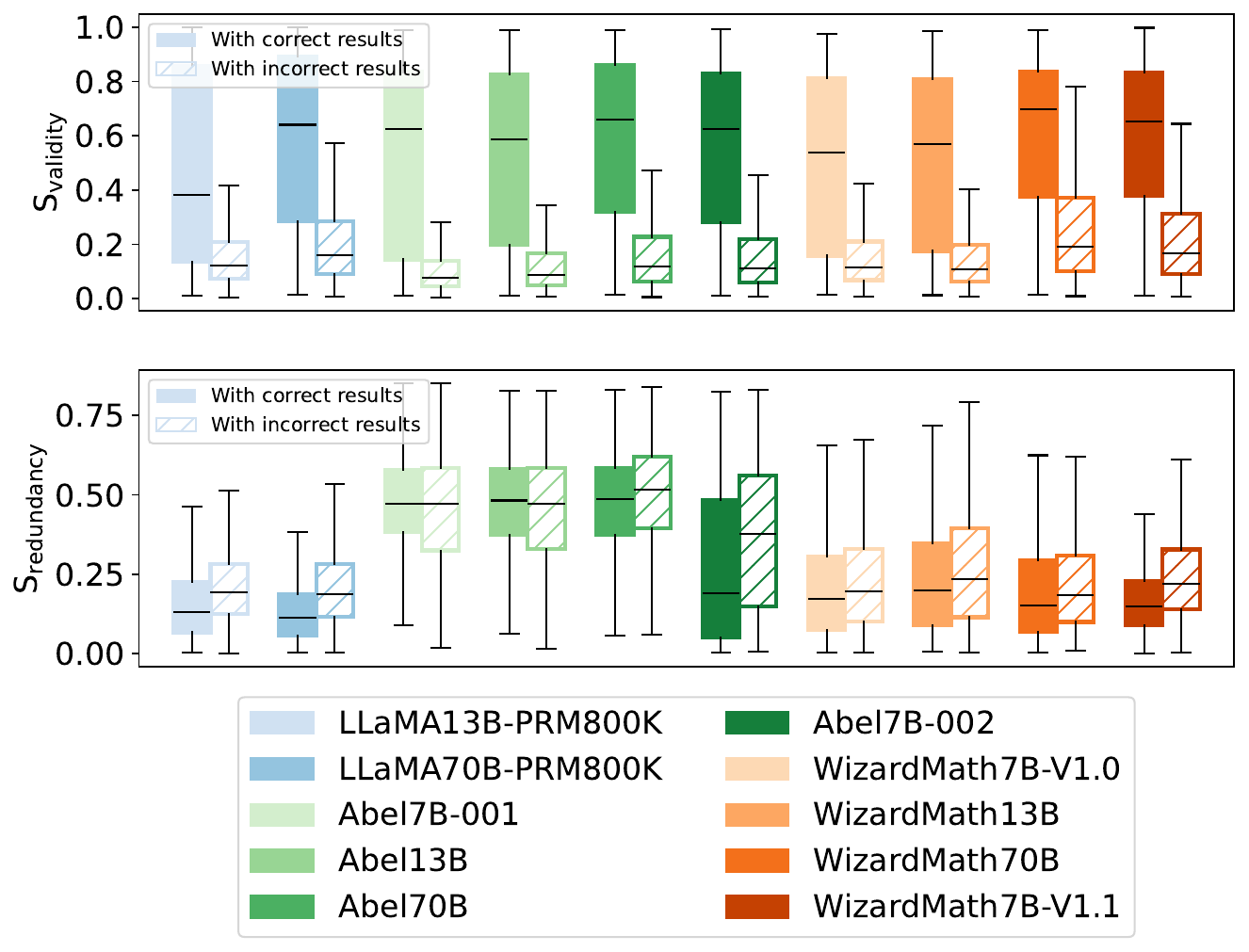}
\caption{Box-and-whisker plots of $S_{\text{validity}}$ (upper) and $S_{\text{redundancy}}$ (lower). The boundaries of the whiskers are based on the 1.5 interquartile range.}
\label{scores_llm}
\end{figure}
\paragraph{Setup} The models selected for measurement are two mainstream LLMs specialized in math: Abel~\citep{abel} and WizardMath~\citep{luo2023wizardmath}, with different scales. We also report the results of LLaMA-2~\citep{touvron2023llama} naive fine-tuned on PRM800K (approximately 6K solutions that reach the correct final answers) for comparison. We evaluate the performance of the LLMs on MATH and sample 100 solutions for each problem to reduce evaluation noise and sampling randomness. Specifically, the sampling temperature is set to 0.6, and the top-p value is set to 0.95. The maximum sample length is set to 2048 tokens. All solutions are scored by \modelname$_{\text{LLemma-34B}}$.

Our analysis includes two aspects: (1) False positive rate: it refers to the proportion of solutions that have the correct final answers but contain incorrect steps among all solutions with correct final answers. (2) The redundancy of solutions.

\subsubsection{False Positive Rate}

\begin{table}[htbp]
  \centering
    \begin{tabular}{lcc}
    \toprule
    \textbf{Model} &   \textbf{Acc. (\%)} & \textbf{FPR (\%)} \\
    \midrule
    LLaMA2-13B-PRM800K &  7.4 & 40.6 \\
    LLaMA2-70B-PRM800K &     14.7 & 21.8 \\
    \midrule
    Abel7B-001 &  12.4 & 31.8 \\
    Abel13B &        15.2 & 29.8 (\faUser29.2) \\
    Abel70B &      25.7 & 20.0 \\
    Abel7B-002 &  30.0 & 22.8 \\
    \midrule
    WizardMath7B-V1.0 &  8.9 & 34.4 \\
    WizardMath13B &      10.9 & 31.3 (\faUser28.3) \\
    WizardMath70B &       18.7 & 16.5 \\
    WizardMath7B-V1.1  & 31.0 & 16.7 \\
    \bottomrule
    \end{tabular}%
      \caption{We estimate the false positive rate (FPR) with \modelname$_{\text{LLemma-34B}}$. We also check the false positive rate of Abel13B and WizardMath13B by human (denoted by \faUser.), sampling one solution for each problem. ``Acc.'' represents the overall accuracy.}
  \label{tab:false_positive_rate}%
\end{table}%

\begin{figure}[htbp]
\centering 
    \includegraphics[width=0.45\textwidth]{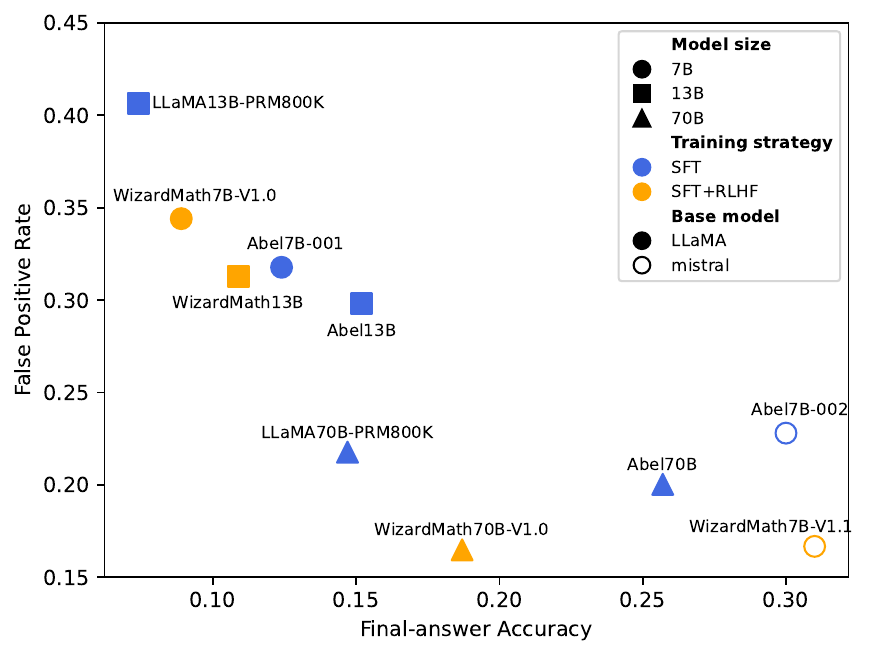}
\caption{Correlation between the final-answer accuracy and the false positive rate.}
\label{estimated false positive}
\end{figure}

We set the threshold to 0.25 for the validity scores and calculate the false positive rate. We describe the details of choosing the threshold value in Appendix \ref{choosing_threshold}. We present the main results in Table \ref{tab:false_positive_rate}. We also include the results in Figure \ref{estimated false positive} with the information of the tested models. We highlight three key takeaways:

\textbf{Increased final-answer accuracy does not inherently ensure a corresponding reduction in the false positive rate.} In Figure \ref{estimated false positive}, it is clear that the false positive rate drops and stays at a level of about 20\% as the final-answer accuracy rises. When comparing WizardMath7B-V1.1 with WizardMath70B, although the former exhibits a much higher accuracy (31.0\% vs. 18.7\%), its false positive rate is almost the same (16.7\% vs. 16.5\%). It indicates there exists a bottleneck for the quality of reasoning steps to advance.

\textbf{The model size and base model affect the false positive rate significantly.} When comparing LLaMA13B-PRM800K with LLaMA70B-PRM800K, although using the same training data, the distinction in false positive rates is large (40.6\% vs. 21.8\%). It indicates the importance of the model scale. The base model of Mistral~\citep{jiang2023mistral} also achieves a lower false positive rate than LLaMA2 (22.8\% vs. 31.8\%, 16.7\% vs. 34.4\%). Overall, the scaling law still applies to this field.

\textbf{RLHF helps lower the false positive rate when the LLMs are relatively strong.} There's no distinction in false positive rates between SFT and SFT plus RLHF when the model is relatively weak, like 7B and the 13B model of LLaMA-2 (31.8\% vs. 34.4\%, 29.8\% vs. 31.3\%). As the model size becomes larger, SFT plus RLHF performs better (16.5\% vs. 20.0\%, 16.7\% vs. 22.8\%). This indicates that effective supervision by RLHF requires a strong mathematical ability of the model itself.

\subsubsection{Redundancy of Solutions}

We analyze the redundancy in reasoning steps. Our main results are in Figure \ref{scores_llm}. It is noteworthy that the redundancy scores of Abel family are usually higher. This is because the solutions from Abel often involve restating the problem first, which is considered a meaningless action towards solving the problem by our evaluator. Moreover, the redundancy scores for solutions that fail to reach the correct answers are higher. This suggests that when a model is unsure about how to solve a problem, it tends to make more attempts that lack meaningful progression.

\subsection{Improving Reasoning Quality by Selecting Training Data}
In this part, we explore the potential of \modelname\ to select high-quality training data for SFT.

\paragraph{Setup}The MMIQC~\citep{liu2024augmenting} dataset consists of a mixture of processed web data and synthetic question-response pairs used to enhance the reasoning capabilities of LLMs. We randomly sample 10K unique responses generated by GPT-3.5-turbo on \texttt{MATH} from the dataset. We then filter it using $\modelname_{\text{WizardMath-7B-V1.1}}$, specifically removing samples with validity scores below 0.5 or redundancy scores above 0.15. We also combine these two conditions. We SFT the mistral-7b in different subsets with 3 epochs and observe the performance on \texttt{MATH}. To compare, we randomly sample the same amount of training data three times and report the average performance.

\paragraph{Results}We make the following observations from Table \ref{filter}: (1) In terms of accuracy, all groups filtered by $\modelname_{\text{WizardMath-7B-V1.1}}$ outperform the random group and achieve performance close to that of the full dataset. (2) The average number of tokens for the solutions decreases in groups filtered by $\modelname_{\text{WizardMath-7B-V1.1}}$, indicating its advantage in increasing problem-solving efficiency. (3) The group that combines the two filtering conditions significantly improves the quality of reasoning steps with only about half the training data.
\begin{table}[htbp]
  \centering
    \begin{tabular}{lcccccc}
        \toprule
   \textbf{Filter}  & \textbf{\#D} & \textbf{Acc.} & \textbf{Val.} & \textbf{Red.} & \textbf{\#Token}\\
\midrule
    -  & 100\% & 22.2  & 65.2  & 27.4  & 723.4  \\
    \midrule
    val.  & 76.7\% &22.0  & 65.9  & 26.4  & 699.9  \\
    random  & 76.7\% & 20.1  & 62.5  & 27.4  & 765.6  \\
    \midrule
    red.  & 71.9\% &21.8  & 65.6  & 22.1  & 681.5  \\
    random  & 71.9\% &20.3  & 62.3  & 28.0  & 746.1  \\
    \midrule
    \makecell[l]{red. \& val.} & 56.7\%  & 22.0  & 67.8  & 22.5  & 701.2  \\
    random & 56.7\% & 20.0  & 62.1  & 27.6  & 739.5  \\
    \bottomrule
    \end{tabular}
      \caption{``\#D'' represents the percentage of training data from the full set. ``Acc.'' represents the overall accuracy. ``Val.'' and ``Red.'' represent $S_{\text{validity}}$ and $S_{\text{redundancy}}$ for the solutions with correct results. ``\#Token'' represents the average number of tokens for the solutions.}
  \label{filter}%
\end{table}

\section{Related Work}
\paragraph{Evaluating reasoning steps automatically} The technique used to assess reasoning steps automatically can be broadly divided into three groups: (1) Embedding-based methods: \citet{golovneva2022roscoe} propose several metrics and uses the embedding-based calculation among hypothesis steps, reference steps, and problems to represent them; (2) Parsing-based methods: this approach aims to parse steps into structured forms, like ‘subject-verb-object’ frames~\citep{prasad2023receval} or symbolic proofs~\citep{saparov2022language}. However, it presents challenges for complex datasets such as MATH due to the intricate logic involved; (3) Prompting-based methods:  due to the generalization of LLMs, given tailored prompts to SOTA LLMs, they can check the solutions and find the potential errors~\citep{tyen2023llms,zeng2024mrgsm8k}.

\paragraph{Process reward model (PRM)} Process reward models are mainly used in reinforcement learning to give feedback to LLMs to align with human logic in mathematics. \citet{lightman2023let} and \citet{uesato2022solving} both evaluate the performance of PRM as a verifier against the outcome reward model. \citet{ma2023let} combine a check-and-generation idea with PRM to generate a more accurate response. \citet{wang2023math} propose a way to automatically construct process-wise supervision data. These studies focus on leveraging PRM to enhance accuracy. In contrast, our research explores the potential of PRM to develop new evaluation methods in mathematics, moving beyond mere accuracy improvements.

\section{Conclusion}
In this work, we develop \modelname, a new metric to evaluate the quality of reasoning steps from the perspectives of correctness and efficiency. We construct a meta-evaluation testbed and show $\modelname_{}$ using the optimal design can consistently outperform baseline methods by a large margin. Additionally, we empirically demonstrate the strong generalization ability of \modelname. With \modelname, we find an inconsistency between final-answer accuracy and the quality of reasoning steps. We also show its efficacy in selecting training data.

\bibliography{aaai25}

\appendix

\section{Ablation Studies on the Scoring Scheme}\label{scoring_scheme}

\begin{table*}[htbp]
  \centering
    \begin{tabular}{lcccc}
    \toprule
          & \multicolumn{2}{c}{Solution-level} & \multicolumn{2}{c}{Step-level} \\
          \cmidrule(lr){2-3} \cmidrule(lr){4-5}
          & $S_{1}$-AUC & $S_{2}$-AUC & $S_{1}$-AUC & $S_{2}$-AUC \\
          \midrule
    \modelname$_{\text{Llama2-7B}}$ & 79.5  & 74.0  & 80.0  & 67.1  \\
    \modelname$_{\text{WizardMath-7B-V1.0}}$ & 81.9  & 80.3  & 83.9  & 77.5  \\
    \modelname$_{\text{Mistral-7B}}$ & 85.1  & 76.5  & 85.7  & 73.5  \\
    \modelname$_{\text{Llemma-7B}}$ & 84.3  & 82.9  & 90.5  & 85.9  \\
    \modelname$_{\text{Abel-7B-002}}$ & 86.2  & 83.5  & 90.5  & 81.3  \\
    \modelname$_{\text{WizardMath-7B-V1.1}}$ & 87.5  & 84.3  & 89.5  & 83.1  \\
    \modelname$_{\text{Llemma-34B}}$ & 90.8  & 89.7  & 92.8  & 89.6  \\
    \bottomrule
    \end{tabular}%
      \caption{Performance of different scoring schemes on MR-MATH-invalid. The $S_1$ represents the scoring scheme implemented by Eq.\ref{s1}, and $S_2$ represents the scoring scheme implemented by Eq.\ref{s2}.}
  \label{ablation_validity_score}%
\end{table*}%
\begin{table*}[htbp]
  \centering
    \begin{tabular}{lcccccc}
    \toprule
          & \multicolumn{3}{c}{MR-MATH-invalid} & \multicolumn{3}{c}{MR-MATH-redundant} \\
          \cmidrule(lr){2-4} \cmidrule(lr){5-7}
          & MIN   & AM    & GM    & MAX   & AM    & GM \\
    \midrule
    \modelname$_{\text{Llama2-7B}}$ & 80.0  & 74.0  & 73.5  & 62.8  & 65.5  & 66.6  \\
    \modelname$_{\text{WizardMath-7B-V1.0}}$& 83.9  & 80.3  & 78.3  & 65.6  & 66.2  & 66.2  \\
    \modelname$_{\text{Mistral-7B}}$ & 85.7  & 76.5  & 80.2  & 63.4  & 64.4  & 64.5  \\
    \modelname$_{\text{Llemma-7B}}$ & 90.5  & 82.9  & 82.5  & 63.0  & 63.5  & 63.2  \\
    \modelname$_{\text{Abel-7B-002}}$ & 90.5  & 83.5  & 85.6  & 63.6  & 64.7  & 65.3  \\
    \modelname$_{\text{WizardMath-7B-V1.1}}$ & 89.5  & 84.3  & 87.9  & 64.8  & 65.4  & 65.4  \\
    \modelname$_{\text{Llemma-34B}}$ & 92.8  & 89.7  & 90.3  & 62.7  & 64.2  & 64.2  \\
    \bottomrule
    \end{tabular}%
   \caption{Performance of different aggregation methods on MR-MATH. The ``AM'' represents the arithmetic mean of the step scores, and ``GM'' represents the geometric mean of the step scores.}
     \label{ablation_aggregation_methods}%
\end{table*}
\paragraph{Validity Scores} The validity score can be defined using two different methods as follows:
\begin{align}
S_{\text{validity}} &= p_{\text{positive}} + p_{\text{neutral}} \tag{1} \label{s1} \\
S^{'}_{\text{validity}} &= p_{\text{positive}} \tag{2} \label{s2}
\end{align}
We compare the performance of the two scoring schemes in Table~\ref{ablation_validity_score}. The scoring scheme Eq.\ref{s1} consistently outperforms Eq.\ref{s2}.

\paragraph{Aggregation Methods} We explore three different methods for aggregating step-level scores: min/max, the arithmetic mean (AM), and the geometric mean (GM). We compare the performance in Table~\ref{ablation_aggregation_methods}. For the validity scores, the min operation consistently outperforms both the AM and GM by a large margin. Conversely, for the redundancy scores, the AM and GM slightly outperform the max operation. This discrepancy may be due to the fact that the redundancy score estimations are less accurate compared to validity scores for \modelname, and averaging operations help to reduce noise. Given the trade-offs and considering the potential future advancements in model capabilities, we ultimately choose the min/max operation, as also taken by \citet{golovneva2022roscoe} and \citet{prasad2023receval}.

\section{Details for \modelname} \label{Training_details}

\paragraph{Training Details}The \modelname\ model is trained on NVIDIA A100 GPUs (4 GPUs for the 7B models, 16 GPUs for the 34B models) using the supervised fine-tuning (SFT) framework provided by~\citet{dai2023safe}. Specifically, the model is trained in 1 epoch. We set the batch size to 64 and maximum sequence length to 2048. We use a peak
learning rate 1e-6 with 8 warmup steps and cosine learning rate decay to 0. We use AdamW~\citep{loshchilov2018decoupled} as our optimizer with $\beta_1 = 0.9, \beta_2 = 0.95$ and weight decay of 0.1.

\paragraph{Splitting Solutions}
For users of \modelname, we suggest splitting solutions into steps using delimiters such as ``\texttt{\textbackslash n}'' or period marks. In our experiments with the Abel and WizardMath series, solutions naturally contained step markers (e.g., ``Step 1:''), allowing for direct step-wise splitting. For the LLaMA-PRM800K series, we split solutions using double newlines (``\texttt{\textbackslash n\textbackslash n}'').

\section{Examples of Solutions Generated by Different LLMs} \label{appendix:lanuguage style}
\begin{figure*}[htbp]
\centering 
    \includegraphics[width=\textwidth]{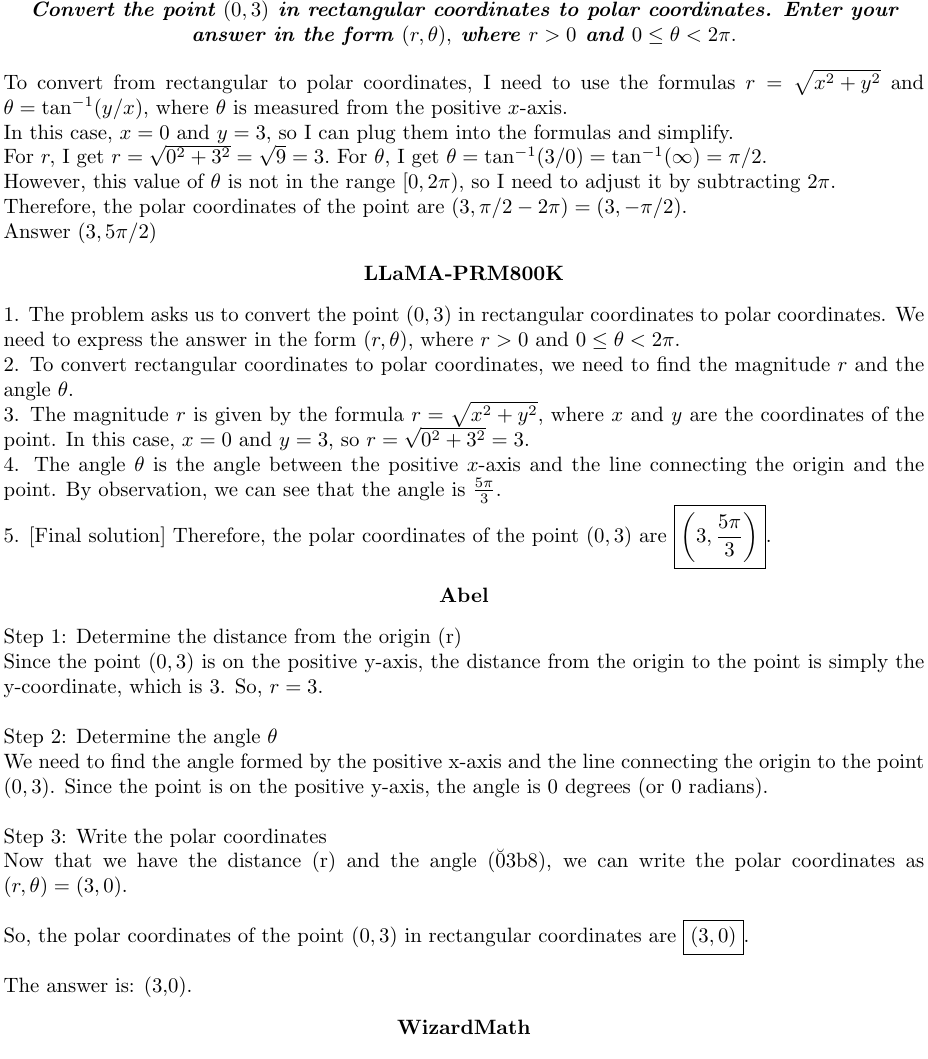}
\caption{Examples of solutions generated by different LLMs}
\label{language style samples}
\end{figure*}

We present examples of solutions generated by different LLMs in Figure \ref{language style samples}, including LLaMA-PRM800K\footnote{It represents the LLaMA-2 naive fine-tuned on PRM800K (approximately 6K solutions that reach the correct final answers), reflecting the language style of PRM800K.}, Abel and WizardMath.

\section{Settings for Evaluators} \label{settings_for_meta_evalautions}
The settings for evaluators are as follows: (1) Embedding-based methods: for ROSCOE we use the embedding layer ``roscoe-512-roberta-base''. We select \textit{Repetition-token} matrix for ROSCOE-SA and \textit{the Repetition-step} matrix for ROSCOE-SS. The threshold is set to 0.025. (2) Prompting-based methods: we present the prompts in Figure \ref{prompt}. (3) Step-level evaluators: for the validity scores, the threshold is set to 0.5, and for redundancy scores, it is set to 0.15. Both are applied to the solution-level labels and step-level labels. For math-shepherd we set the threshold to 0.5 for all errors.
\begin{figure*}[ht!]
\centering
\resizebox{\textwidth}{!}{%
\begin{tabular}{|c|c|}
\hline
\textbf{The Prompt for identifying invalid errors} & \textbf{The Prompt for identifying redundant errors} \\
\hline
\begin{minipage}[t]{0.45\textwidth}

Question:\\
\{question\}\\

Student Solution:\\
\{solution\}\\

Your task involves three parts:\\
1. **Step-by-step Evaluation:** Go through the student solution carefully and identify key errors and potential misunderstandings that led to the incorrect solution.\\
2. **Final Judgement:**  Provide an overall judgement on the correctness of the student's solution.\\
3. **First Error Step:** If the solution is incorrect, generate the step number where the first error occurs, otherwise generate N/A here\\

Here's the format I want:\\
Step-by-step Evaluation: [Provide a step by step examination of the student solution and identify key errors and misunderstandings here.]\\
Final Judgement: [Insert only **correct** or **wrong** here]\\
First Error Step: [Insert either N/A or the step number where the first error occurs]\\

Please follow this format without any additional introductory or concluding statements.
\end{minipage}
&
\begin{minipage}[t]{0.45\textwidth}

Question:\\
\{question\}\\

Student Solution:\\
\{solution\}\\

Your task involves three parts:\\
1. **Step-by-step Evaluation:** Carefully review the student's solution. Identify any neutral steps that, while reasonable, do not offer new insight, advance the solution, or suggest a next step.\\
2. **Final Judgment:** Indicate whether the solution contains or does not contain any neutral steps.\\
3. **Neutral Steps Identified**: If neutral steps are present, list their numbers in list format; otherwise, insert N/A.\\

Here's the format I want:\\
Step-by-step Evaluation: [Provide a step-by-step examination of the student solution and identify the neutral steps.]\\
Final Judgment: [Insert only the word **Present** if neutral steps are identified, or **Absent** if not.]\\
Neutral Steps Identified: [Insert either N/A or the step numbers in list format.]\\

Please follow this format without any additional introductory or concluding statements.
\end{minipage}
\\ \hline
\end{tabular}
}
\caption{Prompts used to identify invalid and redundant errors, modified from the prompt provided by~\citet{zeng2024mrgsm8k}.}
\label{prompt}
\end{figure*}

\section{Threshold Selection for False Positive Rate Estimation} \label{choosing_threshold}
First, we compare the validity scores of reasoning steps from different models in Figure \ref{scores_llm}. For solutions with the correct answers but wrong steps, the validity scores should behave similarly to solutions with the incorrect answers. So we can use it to estimate the threshold and calculate the false positive rate. A rough estimation for the threshold is between 0.15 and 0.35. To further ensure a relatively accurate estimation, we check the false positive rate of Abel13B and WizardMath13B by humans, sampling one solution for each problem. Combining above information, the threshold is set to 0.25.

\end{document}